\documentclass[11pt,a4paper]{article}
\usepackage[hyperref]{acl2021}
\usepackage{times}
\usepackage{latexsym}

\usepackage{graphicx}
\usepackage[color=red!5]{todonotes}

\usepackage{tikz}
\usepackage{wrapfig}
\usepackage{csquotes}
\usepackage{booktabs}

\usepackage{xr}

\makeatletter
\newcommand*{\addFileDependency}[1]{
  \typeout{(#1)}
  \@addtofilelist{#1}
  \IfFileExists{#1}{}{\typeout{No file #1.}}
}
\makeatother

\usepackage{microtype}

\aclfinalcopy 

\title{Supporting Cognitive and Emotional Empathic Writing of Students}

\author{Thiemo Wambsganss\textsuperscript{1,2}, Christina Niklaus\textsuperscript{1, 3}, Matthias Söllner\textsuperscript{4},\\ \textbf{Siegfried Handschuh\textsuperscript{1, 3}} \and \textbf{Jan Marco Leimeister\textsuperscript{1, 4}} \\

  \textsuperscript{1} University of St.Gallen\\
 \footnotesize{ {\tt \{thiemo.wambsganss, christina.niklaus,}}\\ \footnotesize{ {\tt siegfried.handschuh, janmarco.leimeister\}{\tt @unisg.ch}}}\\
  \textsuperscript{2} Carnegie Mellon University\\
  \footnotesize{ \tt {twambsga@andrew.cmu.edu}}\\
  \textsuperscript{3} University of Passau\\
  \footnotesize{ {\tt \{christina.niklaus, siegfried.handschuh\}{\tt@uni-passau.de}}}\\
 \textsuperscript{4} University of Kassel\\
  \footnotesize{ {\tt \{soellner, leimeister\}{\tt@uni-kassel.de}}}\\
\\}  
  
\date{}

\begin{document}
\maketitle
\begin{abstract}
We present an annotation approach to capturing emotional and cognitive empathy in student-written peer reviews on business models in German. We propose an annotation scheme that allows us to model emotional and cognitive empathy scores based on three types of review components. Also, we conducted an annotation study with three annotators based on 92 student essays to evaluate our annotation scheme. The obtained inter-rater agreement of $\alpha$=0.79 for the components and the multi-$\pi$=0.41 for the empathy scores indicate that the proposed annotation scheme successfully guides annotators to a substantial to moderate agreement. Moreover, we trained predictive models to detect the annotated empathy structures and embedded them in an adaptive writing support system for students to receive individual empathy feedback independent of an instructor, time, and location. We evaluated our tool in a peer learning exercise with 58 students and found promising results for perceived empathy skill learning, perceived feedback accuracy, and intention to use. 
Finally, we present our freely available corpus of 500 empathy-annotated, student-written peer reviews on business models and our annotation guidelines to encourage future research on the design and development of empathy support systems.
\end{abstract}


\section{Introduction}
\label{intro}
\begin{figure}[!htb]
 \centering
 \includegraphics[width=0.5\linewidth]{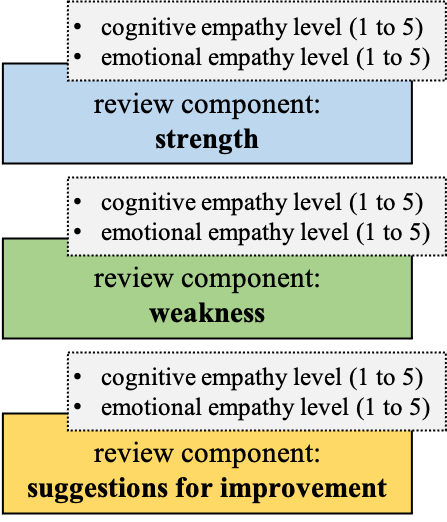}
 \caption{Empathy annotation scheme. First, a text paragraph is classified into a peer review component (\textit{strengths, weakness, improvement suggestions}). Second, the same annotator is then scoring the cognitive and emotional empathy level of the components based on our annotation guideline on a 1-to-5 scale.
} 
 \label{fig:annotationscheme}
\end{figure}
Empathy is an elementary skill in society for daily interaction and professional communication and is therefore elementary for educational curricula (e.g., Learning Framework 2030 \cite{OECD2018The2030}). It is the \textit{``ability to simply understand the other person’s perspective [$\dots$] and to react to the observed experiences of another,''}  \citep[][p.1]{Davis1983MeasuringApproach.}\footnote{Being aware that empathy is a multidimensional construct, in this study, we focus on emotional and cognitive empathy \cite{Spreng2009TheMeasures,Davis1983MeasuringApproach.}.}. Empathy skills not only pave the foundation for successful interactions in digital companies, e.g., in agile work environments \cite{Luca2001DoesTeamwork}, but they are also one of the key abilities in the future that will distinguish the human workforce and artificial intelligence agents from one another \cite{Poser2020HybridAgents}. However, besides the growing importance of empathy, research has shown that empathy skills of US college students decreased from 1979 to 2009 by more than thirty percent and even more rapidly between 2000 to 2009 \cite{Konrath2011ChangesMeta-analysis}. On these grounds, the Organization for Economic Cooperation and Development (OECD) claims that the training for empathy skills should receive a more prominent role in today’s higher education \cite{OECD2018The2030}. To train students with regard to empathy, educational institutions traditionally rely on experiential learning scenarios, such as shadowing, communication skills training, or role playing \cite{Lok2019CanEmpathy,vanBerkhout2016TheTrials}. Individual empathy training is only available for a limited number of students since individual feedback through a student’s learning journey is often hindered due to large-scale lectures or the growing field of distance learning scenarios such as Massive Open Online Classes (MOOCs) \cite{Seaman2018HigherGroup, Hattie2007TheFeedback}. 

One possible path for providing individual learning conditions is to leverage recent developments in computational linguistics. Language-based models enable the development of writing support systems that provide tailored feedback and recommendations \cite{Santos2018AnInteractions}, e.g., like those already used for argumentation skill learning \cite{Wambsganss2020ALSkills,Wambsganss2021ArgueTutor:Skills}. 
Recently, studies have started investigating elaborated models of human emotions (e.g., \newcite{Wang2016DimensionalModel}, \newcite{Abdul-Mageed2017EmoNet:Networks}, \newcite{Buechel2018EmotionLevel}, or \newcite{Sharma2020ASupport}), but available corpora for empathy detection are still rare. 
Only a few studies address the detection and prediction of empathy in natural texts \cite{Khanpour2017IdentifyingCommunities,Xiao2012AnalyzingPsychotherapy.}, and, to the best of our knowledge, only one corpus is publicly available for empathy modelling based on news story reactions \cite{Buechel2018ModelingStories}. Past literature therefore lacks 1) publicly available empathy annotated data sets, 2) empathy annotation models based on rigorous annotation guidelines combined with annotation studies to assess the quality of the data, 3) the alignment of empathy in literature on psychological constructs and theories, and 4) an embedding and real-world evaluation of novel modelling approaches in collaborative learning scenarios \cite{Rose2008AnalyzingLearning}.

We introduce an empathy annotation scheme and a corpus of 500 student-written reviews that are annotated for the three types of review components, \textit{strengths, weaknesses}, and \textit{suggestions for improvements}, and their embedded \textit{emotional} and \textit{cognitive empathy level} based on psychological theory \cite{Davis1983MeasuringApproach.,Spreng2009TheMeasures}. We trained different models and embedded them as feedback algorithms in a novel writing support tool, which provided students with individual empathy feedback and recommendations in peer learning scenarios. 
The measured empathy skill learning \cite{Spreng2009TheMeasures}, the perceived feedback accuracy \cite{Podsakoff1989EffectsPerformance}, and the intention to use \cite{Venkatesh2008TechnologyInterventions} in a controlled evaluation with 58 students provided promising results for using our approach in different peer learning scenarios to offer quality education independent of an instructor, time, and location.

Our contribution is fourfold: 1) we derive a novel annotation scheme for empathy modeling based on psychological theory and previous work on empathy annotation \cite{Buechel2018ModelingStories}; 2) we present an annotation study based on 92 student peer reviews and three annotators to show that the annotation of empathy in student peer reviews is reliably possible; 3) to the best of our knowledge, we present the second freely available corpus for empathy detection in general and the first corpus for empathy detection in the educational domain based on 500 student peer reviews collected in our lecture about business innovation in German; 4) we embedded our annotation approach as predictive models in a writing support system and evaluated it with 58 students in a controlled peer learning scenario. 
We hope to encourage research on student-written empathetic texts and writing support systems to train students' empathy skills based on NLP towards a quality education independent of a student's location or instructors.

\section{Background}
\label{relatedWork}

\paragraph{The Construct of Empathy}
The ability to perceive the feelings of another person and react to their emotions in the right way requires empathy – the ability ``\textit{of one individual to react to the observed experiences of another''} (\newcite{Davis1983MeasuringApproach.}, p.1). Empathy plays an essential role in daily life in many practical situations, such as client communication, leadership, or agile teamwork. Despite the interdisciplinary research interest, the term empathy is defined from multiple perspectives in terms of its dimensions or components \cite{Decety2004TheEmpathy.}. Aware of the multiple perspectives on empathy, in this annotation study, we focused on the cognitive and emotional components of empathy as defined by \newcite{Davis1983MeasuringApproach.} and \newcite{Lawrence2004MeasuringQuotient}. Therefore, we follow the \textit{`Toronto Empathy Scale'} \cite{Spreng2009TheMeasures} as a synthesis of instruments for measuring and validating empathy. 
Hence, empathy consists of both emotional and cognitive components \cite{Spreng2009TheMeasures}. While emotional empathy lets us perceive what other people feel, cognitive empathy is the human ability to recognize and understand other individuals \cite{Lawrence2004MeasuringQuotient}.

\paragraph{Emotion and Empathy Detection}
In NLP, the detection of empathy in texts is usually regarded as a subset of emotion detection, which in turn is often referred to as part of sentiment analysis. The detection of emotions in texts has made major progress, with sentiment analysis being one of the most prominent areas in recent years \cite{Liu2015SentimentEmotions}. However, most scientific studies have been focusing on the prediction of the polarity of words for assessing negative and positive notions (e.g., in online forums \cite{Abbasi2008SentimentForums} or twitter postings \cite{Rosenthal2018SemEval-2017Twitter}). 
Moreover, researchers have also started investigating more elaborated models of human emotions (e.g., \newcite{Wang2016DimensionalModel}, \newcite{Abdul-Mageed2017EmoNet:Networks}, and \newcite{Mohammad2017EmotionTweets}).
Several corpora exist where researchers have annotated and assessed the emotional level of texts. For example, \newcite{Scherer1994EvidencePatterning} published an emotion-labelled corpus based on seven different emotional states.
\newcite{Strapparava2007SemEval-2007Text} classified news headlines based on the basic emotions scale of \newcite{ekman1992argument} (i.e., \textit{anger, disgust,
fear, happiness, sadness} and \textit{surprise}). 
More recently, \newcite{Chen2018EmotionLines:Conversations} published \textit{EmotionLines}, an emotion corpus of multi-party conversations, as the first data set with emotion labels for all utterances was only based on their textual content. 
\newcite{Bostan2018AnGerman} presented a novel unified domain-independent corpus based on eleven emotions as the common label set. 
However, besides the multiple corpora available for emotion detection in texts, corpora for empathy detection are rather rare. As \newcite{Buechel2018ModelingStories} also outline, the construction of corpora for empathy detection and empathy modelling might be less investigated due to various psychological perspectives on the construct of empathy. Most of the works for empathy detection focus, therefore, on spoken dialogue, addressing conversational agents, psychological interventions, or call center applications (e.g., \newcite{McQuiggan2007ModelingAgents}, \newcite{Perez-Rosas2017UnderstandingTherapy}, \newcite{Alam2018AnnotatingConversations}, \newcite{Sharma2020ASupport}) rather than written texts. Consequently, there are hardly any corpora available in different domains and languages that enable researchers in training models to detect the empathy level in texts, e.g., by providing students with individual empathy feedback \cite{Buechel2018ModelingStories}.

\paragraph{Empathy Annotated Corpora and Annotation Schemes} 
Only a few studies address the detection and prediction of empathy in natural language texts (e.g., \newcite{Khanpour2017IdentifyingCommunities} and \newcite{Xiao2012AnalyzingPsychotherapy.}). Presenting the first and only available gold standard data set for empathy detection, \newcite{Buechel2018ModelingStories} constructed a corpus in which crowdworkers 
were asked to write emphatic reactions to news stories. Before the writing tasks, the crowdworkers were asked to conduct a short survey with self-reported items to measure their empathy level and their personal distress based on \newcite{Batson1987DistressConsequences}. The scores from the survey were then taken as the annotation score for the overall news reaction message. The final corpus consisted of 1,860 annotated messages \cite{Buechel2018ModelingStories}.  
Nevertheless, previous empathy annotations on natural texts merely focused on intuition-based labels instead of rigorous annotation guidelines combined with annotation studies by researchers to assess the quality of the corpora (i.e., as is done for corpora of other writing support tasks, e.g., argumentative student essays by \newcite{Stab2017ParsingEssays}). Moreover, previous annotations have mostly been conducted at the overall document level, resulting in one generic score for the whole document, which makes the corpus harder to apply to writing support systems. 

Consequently, there is a \textit{lack of linguistic corpora for empathy detection in general} and, more specifically, for training models that provide students with adaptive support and feedback about their empathy in common pedagogical scenarios like large-scale lectures or the growing field of MOOCs \cite{wambsganssempathy, wambsganss-etal-2020-corpus}.
In fact, in the literature about computer-supported collaborative learning 
\cite{Dillenbourg2009}, we found only one approach by \citet{Santos2018AnInteractions} that used a dictionary-based approach to provide students with feedback on the empathy level of their texts. We aim to address this literature gap by presenting and evaluating an annotation scheme and an annotated empathy corpus built on student-written texts with the objective to develop intelligent and accurate empathy writing support systems for students. 

\section{Corpus Construction}
\label{annotationScheme}
We compiled a corpus of 500 student-generated peer reviews in which students provided each other with feedback on previously developed business models. Peer reviews are a modern learning scenario in large-scale lectures, enabling students to reflect on their content, receive individual feedback from peers, and thus deepen their understanding of the content \cite{Rietsche2019InsightsSkill}. Moreover, they are easy to set up in traditional large-scale learning scenarios or the growing field of distance-learning scenarios such as MOOCs. This can be leveraged to train skills such as \textit{the ability to appropriately react to other students’ perspectives} (e.g., \newcite{Santos2018AnInteractions}). Therefore, we aim to create an annotated corpus to provide empathy feedback based on a data set that A) is based on real-world student peer reviews, B) consists of a sufficient corpus size to be able to train models in a real-world scenario and C) follows a novel annotation guideline for guiding the annotators towards an adequate agreement. Hence, we propose a new annotation scheme to model peer review components and their emotional and cognitive empathy levels that reflect the feedback discourse in peer review texts. We base our empathy annotation scheme on emotional and cognitive empathy following \newcite{Davis1983MeasuringApproach.} and \newcite{Spreng2009TheMeasures} guided by the study of \newcite{Buechel2018ModelingStories}. To build a reliable corpus, we followed a 4-step methodology: 1) we examined scientific literature and theory on the construct of empathy and on how to model empathy structures in texts from different domains; 2) we randomly sampled 92 student-generated peer reviews and, on the basis of our findings from literature and theory, developed a set of annotation guidelines consisting of rules and limitations on how to annotate emphatic review discourse structures; 3) we applied, evaluated, and improved our guidelines with three native speakers of German in a total of eight consecutive workshops to resolve annotation ambiguities; 4) we followed the final annotation scheme based on our 14-page guidelines to annotate a corpus of 500 student-generated peer reviews.\footnote{The annotation guidelines as well as the entire corpus can be accessed at \url{https://github.com/thiemowa/empathy_annotated_peer_reviews}.} 


\subsection{Data Source}
\label{Data}
We gathered a corpus of 500 student-generated peer reviews written in German. The data was collected in a business innovation lecture in a master's program at a Western European university. In this lecture, around 200 students develop and present a new business model for which they receive three peer reviews each. Here, a fellow student from the same course elaborates on the strengths and weaknesses of the business model and gives recommendations on what could be improved. We collected a random subset of 500 of these reviews from around 7,000 documents collected from the years 2014 to 2018 in line with the ethical guidelines of our university and with approval from the students to utilize the writings for scientific purposes. An average peer review consists of 200 to 300 tokens (in our corpus we counted a mean of 19 sentences and 254 tokens per document). A peer review example is displayed in Figure \ref{fig:example}.

\subsection{Annotation Scheme}
\label{annotation}
Our objective is to model the empathy structures of student-generated peer reviews by annotating the review components and their emotional and cognitive empathy levels. Most of the peer reviews in our corpus followed a similar structure. They described several strengths or weaknesses of the business model under consideration, backing them up by examples or further elaboration. Moreover, the students formulated certain suggestions for improvements of the business model. These review components (i.e., \textit{strengths, weaknesses,  and suggestions for improvement}) were written with different empathetic levels, sometimes directly criticizing the content harshly, sometimes empathetically referring to weaknesses as further potentials for improvement with examples and explanation. We aim to capture these empathic differences between the peer reviews with two empathy level scores, the \textit{cognitive empathy level} of a certain review component and the \textit{emotional empathy} level of a certain component. Our basic annotation scheme is illustrated in Figure \ref{fig:annotationscheme}.

\subsubsection{Review Components}
\label{reviewcomponent}
For the review components, we follow established models of feedback structures suggested by feedback theory (e.g., \newcite{Hattie2007TheFeedback} or \newcite{Black2009DevelopingAssessment}). A typical peer review, therefore, consists of three parts: 1) elaboration of strengths, 2) elaboration of weaknesses, and 3) suggestions for improvements (to answer \textit{``Where am I going and how am I going?''} and \textit{``Where do I go next?''}, i.e., \newcite{Hattie2007TheFeedback}). Accordingly, the content of a review consists of multiple components, including several controversial statements (e.g., a claim about a strength or weakness of a business model) that are usually supported by elaborations or examples (i.e., a \textit{premise}) \cite{Toulmin1984IntroductionReasoning}. Also, in the domain of student-written peer reviews, we found that a standpoint and its elaboration are the central element of a review component. 
Accordingly, we summarized all the claims and premises which described positive aspects of a business model as \textit{strengths}. All content (claims and premises) describing negative aspects were modelled as \textit{weaknesses}, while claims and premises with certain content for improvement were modelled as \textit{suggestions for improvement}, following the structure of a typical review. Besides the content, syntactical elements and key words were used as characteristics for the compound classification, e.g., most students introduced a review component by starting with structural indications such as \textit{"Strengths:"} or \textit{"Weaknesses:" }in their peer review texts. 

\subsubsection{Empathy Level}
\label{empathylevel}

To capture the differences in the empathy levels of the peer reviews (i.e., the way the writer was conveying their feedback \cite{Hattie2007TheFeedback}), we followed the approach of \newcite{Davis1983MeasuringApproach.} and \newcite{Spreng2009TheMeasures} for cognitive and emotional empathy.
\begin{table*}[!htb]
\centering
\footnotesize
\begin{tabular}{p{0.3cm}p{14.9cm}}
\toprule
 Score & Description \\ \hline\hline
 5 & The student fully understands the peer’s thoughts. She completely stepped outside her own perspective and thinks from the peer’s perspective. She does that by carefully evaluating the peer’s idea with rich explanations. Questions, personal pronouns, or direct addressing of the author could be used in order to better understand and elaborate on the peer’s perspective. \\ \hline
 4 & The student thinks from the perspective of the peer. She elaborates in a way that serves the peer best to further establish the idea or activity. Each component is affirmed with further explanations. \\ \hline
 3 & The student tries to understand the perspective of the peer and adds further elaborations to her statements. However, her elaborations are not completely thought through, and her feedback is missing some essential explanations, examples, or questions to make sure she understood everything correctly. \\ \hline
 2 & The student did not try to understand the peer’s perspective. The student rather just tried to accomplish the task of giving feedback.\\ \hline
 1 & The student’s feedback is very short and does not include the peer’s perspective. She does not add any further elaboration in her thoughts. \\ \bottomrule
\end{tabular}
\caption{Description of the cognitive empathy scores.}
\label{tab:cognitivEmpathyScore}
\end{table*}
Cognitive empathy (perspective taking) is the writer's ability to use cognitive processes, such as role taking, perspective taking, or “\textit{decentering},” while evaluating the peers’ submitted tasks. The student sets aside their own perspective and \textit{“steps into the shoes of the other.”} Cognitive empathy can happen purely cognitively, in that there is no reference to any affective state, \cite{Baron-Cohen2004TheDifferences} but it mostly includes understanding the other’s emotional state as well. The following example displays high cognitive empathy: \textit{``You could then say, for example, ‘Since market services are not differentiated according to customer segments and locations, the following business areas result... And that due to the given scope of this task you will focus on the Concierge-Service business segment.’ After that, you have correctly only dealt with this business segment.''}
Emotional empathy (emphatic concern) is the writer's emotional response to the peers’ affective state. The students can either show the same emotions as read in the review or simply state an appropriate feeling towards the peer. Typical examples include sharing excitement with the peer about the business model submitted or showing concern over the peer’s opinion. The following example depicts high emotional empathy: \textit{``I think your idea is brilliant!''}.

Both constructs are measured on a scale from 1-5 following the empathy scale range of \newcite{Moyers2010Revised3.1.1}, with every level being precisely defined in our annotation guidelines. A summary of the definitions for both empathy level scores are displayed in Table \ref{tab:cognitivEmpathyScore} and Table \ref{tab:emotionalEmpathyScore}. A more detailed description of both scores can be found in the appendix in Table \ref{tab:detailedcognitivEmpathyScore} and Table \ref{tab:detailedemotionalEmpathyScore}.\footnote{More elaborated definitions, examples, and key word lists for both empathy scales can be found in our annotation guidelines.}

\begin{table*}[!htb]
\centering
\footnotesize
\begin{tabular}{p{0.3cm}p{14.9cm}}
\toprule
 Score & Description \\ \hline\hline
 5 & The student was able to respond very emotionally to the peer’s work and fully represents the affectional state in her entire review. She illustrates this by writing in a very emotional and personal manner and expressing her feelings (positive or negative) throughout the review. Strong expressions include exclamation marks (!). 
 \\ \hline
 4 & The student was able to respond emotionally to the peer’s submitted activity with suitable emotions (positive or negative). She returns emotions in her feedback on various locations and expresses her feelings by using the personal pronouns (“I”, “You”). Some sentences might include exclamations marks (!). 
 \\ \hline
 3 & The student occasionally includes emotions or personal emotional statements in the peer review. They could be quite strong. However, the student’s review is missing personal pronouns (“I”, “You”) and is mostly written in third person. Emotions can both be positive or negative. Negative emotions can be demonstrated with concern, missing understanding or insecurity (e.g., with modal verbs or words such as rather, perhaps). 
 \\ \hline
 2 & Mostly, the student does not respond emotionally to the peer’s work. Only very minor and weak emotions or personal emotional statements are integrated. The student writes mostly objectively (e.g., “Okay”, “This should be added”, “The task was done correctly”, etc.). In comparison to level 1, she might be using modal verbs (might, could, etc.) or words to show insecurity in her feedback (rather, maybe, possibly). 
 \\ \hline
 1 & The student does not respond emotionally to the peer’s work at all. She does not show her feelings towards the peer and writes objectively (e.g., no “I feel”, “Personally” “I find this...” and no emotions such as “good”, “great”, “fantastic”, “concerned”, etc.). Typical examples would be “Add a picture.” or “The value gap XY is missing.”. 
 \\ \bottomrule
\end{tabular}
\caption{Description of the emotional empathy scores.}
\label{tab:emotionalEmpathyScore}
\end{table*}

\begin{figure}[!htb]
 \centering
 \includegraphics[width=0.8\linewidth]{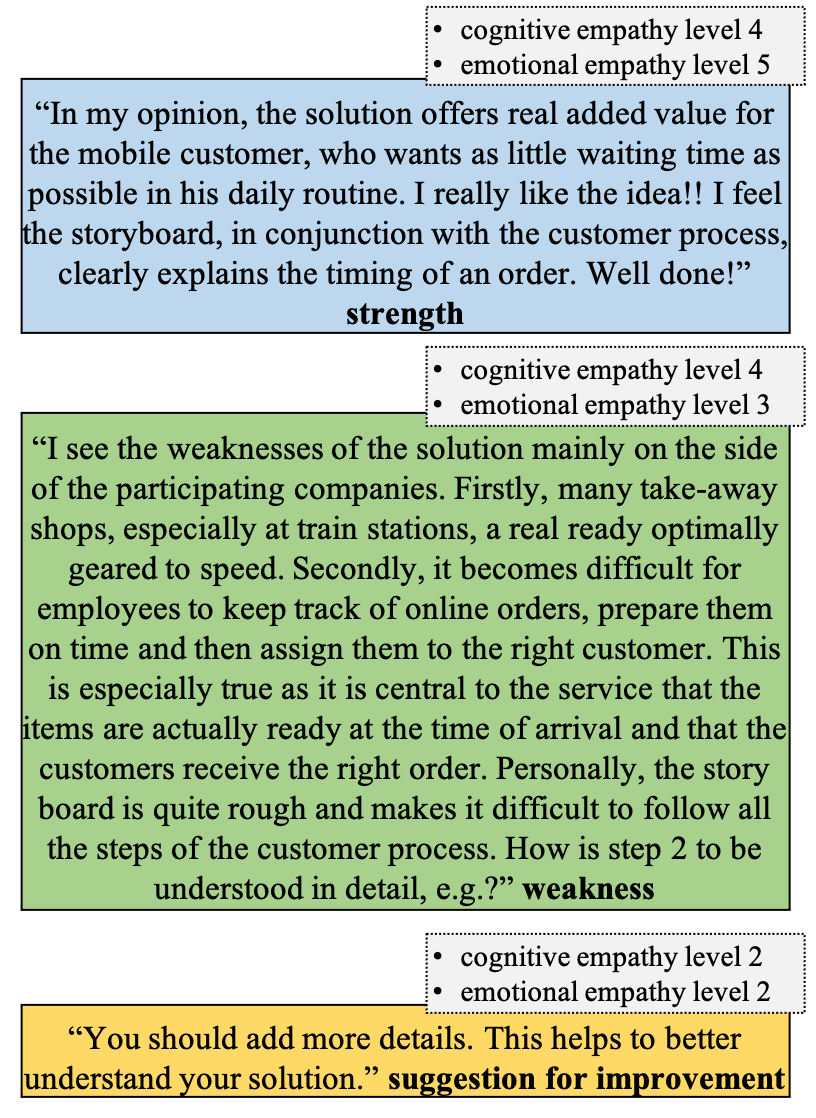}
 \caption{Fully annotated example of a peer review.}
 \label{fig:example}
\end{figure}

Figure \ref{fig:example} illustrates an example of an entire peer review that is annotated for \textit{strength, weakness and suggestion for improvement} and the cognitive and emotional empathy scores.\footnote{Since the original texts are written in German, we translated the examples to English for the sake of this paper.}

\subsection{Annotation Process}

\label{annotationProcess}
Three native German speakers annotated the peer reviews independently from each other for the components \textit{strengths, weaknesses} and \textit{suggestions for improvement}, as well as their \textit{cognitive and emotional empathy levels} according to the annotation guidelines we specified. The annotators were master's students in business innovation from a European university with bachelor's degrees in business administration and were, therefore, domain experts in the field of business models. Inspired by \newcite{Stab2017ParsingEssays}, our guidelines consisted of 14 pages, including definitions and rules for how the review components should be composed, which annotation scheme was to be used, and how the cognitive and emotional empathy level were to be judged. Several individual training sessions and eight team workshops were performed to resolve disagreements among the annotators and to reach a common understanding of the annotation guidelines on the cognitive and emotional empathy structures. We used the \textit{tagtog} annotation tool,\footnote{\url{https://tagtog.net/}} which 
offers an environment for cloud-based annotation in a team. 
First, a text was classified into peer review components (\textit{strengths, weaknesses, suggestions for improvement}, or \textit{none}) by the trained annotators. Second, the same annotator then scored the cognitive and emotional empathy levels of each component based on our annotation guideline on a one to five scale. 
After the first 92 reviews were annotated by all three annotators, we calculated the inter-annotator agreement (IAA) scores (see Section \ref{sec:agreement}).\footnote{Our intention was to capture the annotation of 100 randomly selected essays. However, we discarded 8 of the 100 
 essays as they contained less than 2 review components.} 
 As we obtained satisfying results, we proceeded with two annotators annotating 130 remaining documents each and the senior annotator annotating 148 peer reviews, resulting in 408 additional annotated documents. Together with the 92 annotations of the annotation study of the senior annotator (the annotator with the most reviewing experience), 
 we counted 500 annotated documents in our final corpus.

\section{Corpus Analysis}
\label{corpusAnalysis}

\subsection{Inter-Annotator Agreement}
\label{sec:agreement}
To evaluate the reliability of the review components and empathy level annotations, we followed the approach of \newcite{Stab2014AnnotatingEssays}. 

\paragraph{Review Components}
Concerning the review components, two strategies were used. Since there were no predefined markables, annotators not only had to identify the \textit{type of review component} but also its \textit{boundaries}. In order to assess the latter, we use Krippendorff's $\alpha\textsubscript{U}$ \cite{krippendorff2004}, which allows for an assessment of the reliability of an annotated corpus, considering the differences in the markable boundaries. To evaluate the annotators' agreement in terms of the selected category of a review component for a given sentence, we calculated the percentage agreement and two chance-corrected measures, multi-$\pi$ \cite{fleiss1971mns} and Krippendorff's $\alpha$ \cite{krippendorff80}. Since each annotation always covered a full sentence (or a sequence of sentences), we operated at the sentence level for calculating the reliability of the annotations in terms of the IAA.


\begin{table}[!htb]
\centering
\scriptsize
\begin{tabular}{c|cccc}
\toprule
 & \% & Multi-$\pi$ & Kripp. $\alpha$ & Kripp. $\alpha\textsubscript{U}$\\ \hline
 \textbf{Strength} & 0.9641 & 0.8871 & 0.8871 & 0.5181 \\ 
 \textbf{Weakness} & 0.8893 & 0.7434 & 0.7434 & 0.3109 \\ 
 \textbf{Suggestions} & 0.8948 & 0.6875 & 0.6875 & 0.3512\\ 
 \textbf{None} & 0.9330 & 0.8312 & 0.8312 & 0.9032 \\ \bottomrule
\end{tabular}
\caption{IAA of review component annotations.}
\label{tab:agreement_components}
\end{table}

Table \ref{tab:agreement_components} displays the resulting IAA scores. 
The obtained scores for Krippendorff's $\alpha$ indicated an almost perfect agreement for the \textit{strengths} components and a substantial agreement for both the \textit{weaknesses} and the \textit{suggestions for improvement} components. 
The unitized $\alpha$ of strengths, weaknesses and suggestions for improvement annotations was slightly smaller compared to the sentence-level agreement. Thus, the boundaries of review components were less precisely identified in comparison to the classification into review components. Yet the scores still suggest that there was a moderate level of agreement between the annotators for the strengths and a fair agreement for the weaknesses and the suggestions for improvement. With a score of $\alpha\textsubscript{U}$=90.32\%, the boundaries of the non-annotated text units were more reliably detected, indicating an almost perfect agreement between the annotators. Percentage agreement, multi-$\pi$, and Krippendorff's $\alpha$ were considerably higher for the non-annotated spans as compared to the strengths, weaknesses, and suggestions for improvement
, indicating an almost perfect agreement between the annotators. Hence, we conclude that the annotation of the review components in student-written peer reviews is reliably possible .

\paragraph{Empathy Level}


To assess the reliability of the cognitive and emotional empathy level annotations, we calculated the multi-$\pi$ for both scales. For the cognitive empathy level, we received a multi-$\pi$ of 0.41 for both the emotional and cognitive empathy level, 
suggesting a moderate agreement between the annotators in both cases. 
Thus, we conclude that the empathy level can also be reliably annotated in student-generated peer reviews. 

To analyze the disagreement between the three annotators, we created a confusion probability matrix (CPM) \cite{cinkova-etal-2012-managing} for the review components and the empathy level scores. The results can be found in Section \ref{app:disagreement} of the appendix.

\subsection{Corpus Statistics}


The corpus we compiled consists of 500 student-written peer reviews in German that were composed of 9,614 sentences with 126,887 tokens in total. Hence, on average, each document had 19 sentences and 254 tokens. A total of 2,107 strengths, 3,505 weaknesses and 2,140 suggestions for improvement were annotated.  

Tables \ref{tab:distribution_sentences_tokens}, \ref{tab:distribution_components}, and \ref{tab:distribution_relations} present some detailed statistics on the final corpus. 

\begin{table}[!htb]
\centering
\scriptsize
\begin{tabular}{c|cccccc}
\toprule
 & total & mean & std dev & min & max & median \\ \hline
 \textbf{Sentences} & 9,614 & 19.23 & 10.39 & 1 & 85 & 17 \\ 
 \textbf{Tokens}    & 126,887 & 253.77 & 134.18 & 10 & 1026 & 228 \\ \bottomrule
\end{tabular}
\caption{Distribution of \textit{sentences} and \textit{tokens} in the created corpus. Mean, std dev, min, max and median refer to the number of sentences and tokens per document.}
\label{tab:distribution_sentences_tokens}
\end{table}

\begin{table}[!htb]
\centering
\scriptsize
\begin{tabular}{c|ccccccc}
\toprule
 & total & mean & std dev & min & max & median & \%\\ \hline
 \textbf{Str.} & 2,107 & 4.21 & 2.71 & 1 & 20 & 4 & 0.27 \\ 
 \textbf{Weak.} & 3,505 & 7.01 & 6.10 & 0 & 41 & 5 & 0.45\\ 
 \textbf{Sug.} & 2,140 & 4.28 & 5.49 & 0 & 59 & 3 & 0.28 \\ \bottomrule
\end{tabular}
\caption{Distribution of the 
\textit{review components}.} 
\label{tab:distribution_components}
\end{table}

\begin{table}[!htb]
\centering
\scriptsize
\begin{tabular}{p{2cm}|cccccc}
\toprule
  & mean & std dev & min & max & median \\ \hline
 \textbf{Cognitive EL} & 2.94 & 0.99 & 1 & 5 & 3     \\ 
 \textbf{Emotional EL} & 3.22 & 1.03 & 1 & 5 & 3     \\ \bottomrule 
\end{tabular}
\caption{Distribution of the \textit{empathy level (EL) scores}.}
\label{tab:distribution_relations}
\end{table}

\begin{figure*}[!htb]
\centering
\begin{minipage}[t]{0.43\textwidth}
\includegraphics[width=\textwidth]{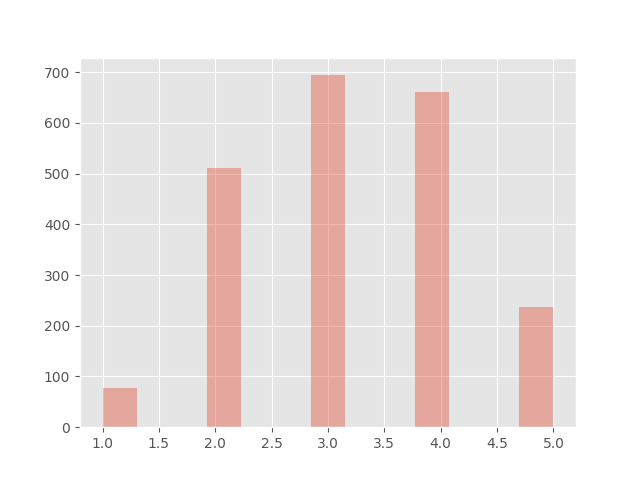}

\end{minipage}
\begin{minipage}[t]{0.43\textwidth}
\includegraphics[width=\textwidth]{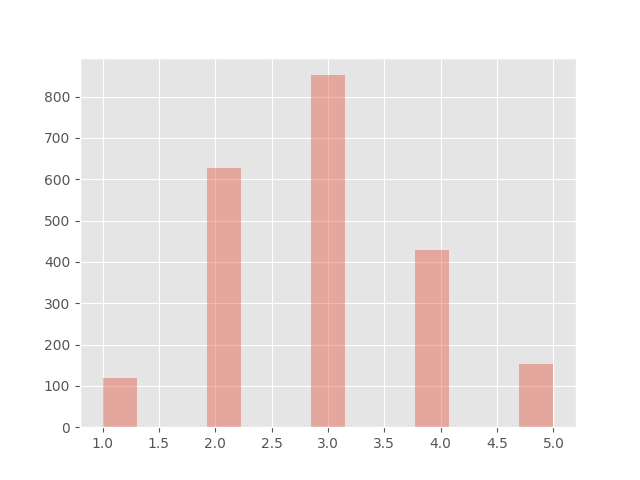}

\end{minipage}
\caption{Distribution of the cognitive \textit{(left)} and emotional \textit{(right)} empathy scores (1-5 scale).}
\label{fig:distribution_empathy_scores}
\end{figure*}

Moreover, Figure \ref{fig:distribution_empathy_scores} displays the distribution of the empathy scores in the annotated dataset. Both the cognitive and the emotional empathy levels approximately follow a normal distribution with a mean score of 2.94 and 3.22, respectively (see Table \ref{tab:distribution_relations}). 
We measured only a low correlation of 0.38 between the scores of cognitive and emotional empathy. 

\section{Providing Students Adaptive Feedback}
\begin{figure*}[!htb]
\centering
\frame{\includegraphics[width=0.95\textwidth]{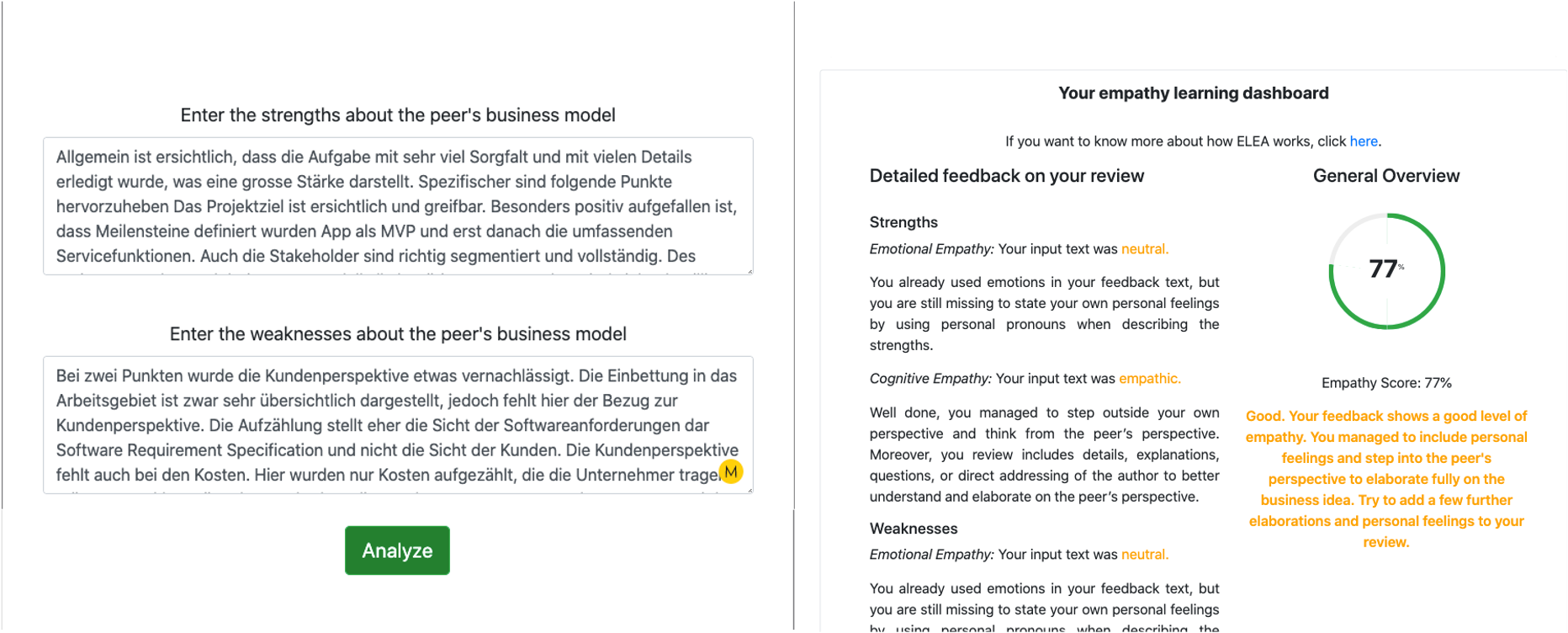}}
\caption{Screenshot of a trained model on our corpus as an adaptive writing support system.}
 \label{fig:application}
\end{figure*}
\paragraph{Modelling Cognitive and Emotional Empathy} 
The empathy detection task is considered a paragraph-based, multi-class classification task, where each paragraph is either considered to be a \textit{strength, weakness}, or a \textit{suggestion for improvement} and has a ``non-empathic'', ``neutral'', or ``empathic'' cognitive and emotional empathy level. Therefore, we assigned the levels of our cognitive and emotional empathy scores to three different labels: level 1 and 2 were assigned to a ``non-empathic'' text label, level 3 to a ``neutral'' label, and levels 4 and 5 to an``empathic'' label . We split the data into 70\% training, 20\% validation, and 10\% test data. To apply the model, the corpus texts were split into word tokens. The model performances were measured in terms of accuracy, precision, recall, and f1-score.

We trained a predictive model following the architecture of Bidirectional Encoder Representations from Transformers (BERT) proposed by \citet{Devlin2018BERT:Understanding}. We used the BERT model from \textit{deepset},\footnote{{\url{https://github.com/deepset-ai/FARM}}} since it is available in German and provides a deep pretrained model that was unsupervised while training on domain-agnostic German corpora (e.g., the German Wikipedia). 
The best performing paramenter combination for our BERT model incorporated a dropout probability of 10\% and a learning rate of 3e\textsuperscript{-5}, and the number of epochs were 3.
After several iterations, we reached a micro f1-score of 74.96\% for the detection of the emotional empathy level and 69.98\% for the detection of the cognitive empathy level of a text paragraph. Moreover, we reached an f1-score of 94.83\% to predict a text paragraph as a strength, a 64.28\% to predict a text paragraph as a weakness, and 59.79\% to predict suggestions for improvement. 
To ensure the validity of our BERT model, we benchmarked against bidirectional Long-Short-Term-Memory-Conditional-Random-Fields classifiers (BiLSTM-CRF). In combination with the corresponding embeddings vocabulary (GloVe) \cite{Pennington2014GloVe:Representation}, our LSTM reached an unsatisfying f1-score of 61\% for detecting the emotional empathy level and 51\% for detecting the cognitive empathy level.

\paragraph{Evaluation in a Peer Learning Setting} 
We designed and built an adaptive writing support system that provides students with individual feedback on their cognitive and emotional empathy skills. The application is illustrated in Figure \ref{fig:application}. We embedded our system into a peer writing exercise where students were asked to write a peer review on a business model. During this writing task, they received adaptive feedback on the cognitive and emotional empathy level based on our model. The evaluation was conducted as a web experiment facilitated by the behavioral lab of our university, and thus, designed and reviewed according to the ethical guidelines of the lab and the university. We received 58 valid results (mean age = 23.89, SD= 3.07, 30 were male, 28 female). The participants were told to read an essay about a business model of a peer student. Afterwards, they were asked to write a business model review for the peer by providing feedback on the strengths, weaknesses, and suggestions for improvement of the particular business model. After the treatment, we measured the intention to use (ITU) \cite{Venkatesh2008TechnologyInterventions} by asking three items. We also asked the participants to judge their perceived empathy skill learning (PESL) by asking two items that covered cognitive and emotional empathy skills \cite{Spreng2009TheMeasures, Davis1983MeasuringApproach.}. Finally, we surveyed the perceived feedback accuracy (PFA) \cite{Podsakoff1989EffectsPerformance} to control the accuracy of our model. All constructs were measured with a 1-to-7 point Likert scale (1: totally disagree to 7: totally agree, with 4 being a neutral statement).\footnote{The exact items are listed in the appendix.} Furthermore, we asked three qualitative questions: ``\textit{What did you particularly like about the use of the tool?}'', ``\textit{What else could be improved?}'', and ``\textit{Do you have any other ideas?}'' and captured the demographics. In total, we asked 13 questions. All participants were compensated with an equivalent of about 12 USD for a 25 to 30 minute experiment.

\paragraph{Results} Participants judged their empathy skill learning with a mean of 5.03 (SD= 1.05). Concerning the PFA, the subjects rated the construct with a mean of 4.93 (SD= 0.94).
The mean value of intention to use of the participants using our application as a writing support tool in peer learning scenarios was 5.14 (SD= 1.14). The mean values of all three constructs were very promising when comparing the results to the midpoints. All results were better than the neutral value of 4, indicating a positive evaluation of our application for peer learning tasks.
We also asked open questions in our survey to receive the participants' opinions about the tool they used. The general attitude was very positive. Participants positively mentioned the simple and easy interaction, the distinction between cognitive and emotional empathy feedback, and the overall empathy score together with the adaptive feedback message several times. However, participants also said that the tool should provide even more detailed feedback based on more categories and should provide concrete text examples on how to improve their empathy score. We translated the responses from German and clustered the most representative responses in Table \ref{tab:qualitativeuser} in the appendix.

\section{Conclusion}
\label{conclusion}


We introduce a novel empathy annotation scheme and an annotated corpus of student-written peer reviews extracted from a real-world learning scenario. 
Our corpus consisted of 500 student-written peer reviews that were annotated for review components and their emotional and cognitive empathy levels. Our contribution is threefold: 1) we derived a novel annotation scheme for empathy modeling based on psychological theory and previous work for empathy modeling \cite{Buechel2018ModelingStories}; 2) we present an annotation study based on 92 student peer reviews and three annotators to show that the annotation of empathy in student peer reviews is reliably possible ; and 3) to the best of our knowledge, we present the second freely available corpus for empathy detection and the first corpus for empathy detection in the educational domain based on 500 student peer reviews in German. For future research, this corpus could be leveraged to support students' learning processes, e.g., through a conversational interaction \cite{Zierau2020TheClues}. However, we would also encourage research on the ethical considerations of empathy detection models in user-based research (i.e., \citet{Wambsganss2021EthicalDesign}).
We, therefore, hope to encourage future research on student-generated empathetic texts and on writing support systems to train empathy skills of students based on NLP towards quality education independent of a student's location or instructors.

\bibliographystyle{acl_natbib}
\bibliography{anthology,acl2021,references.bib,refs.bib}

\appendix

\section{Details on the Description of the Annotation Scheme\footnote{Further examples and descriptions can be found in our annotation guideline.}}

A more detailed description of the cognitive and emotional empathy scores can be found in Table \ref{tab:detailedcognitivEmpathyScore} and Table \ref{tab:detailedemotionalEmpathyScore}. 

\section{Details on the Annotation Process}
The annotation process was split into three steps: 

\begin{enumerate}
\item \textbf{Reading of the entire peer review:} The annotators are confronted with the student-written peer review and are asked to read the whole document. This helps to get a first impression of the review and get an overview of the single components and the structure of it.
\item \textbf{Labeling the components and elaborations:} After reading the entire student-written peer review, the annotator is asked to label the three different components (\textit{strengths, weaknesses} and \textit{suggestions for improvement}). Every supporting sentence (such as explanation, example, etc.) is annotated together with the referred component.
\item \textbf{Classification of the cognitive and emotional empathy levels: }Each component is assessed on its level of cognitive and emotional empathy by giving a number between 1-5. Each category is carefully defined and delimited according to Table \ref{tab:detailedcognitivEmpathyScore} and Table \ref{tab:detailedemotionalEmpathyScore}.
\end{enumerate}

\begin{table*}[!htb]
\centering
\footnotesize
\begin{tabular}{c|p{12cm}}
\toprule
 Score & Description \\ \hline\hline
 5 = strong & The student fully understands the peer’s thoughts. She completely steps outside her own perspective and thinks from the peer’s perspective. She does that by carefully evaluating the peer’s idea with rich explanations. Questions, personal pronouns, or direct addressing of the author can be used in order to better understand and elaborate on the peer’s perspective. 
 
\textit{Strengths: }The student fully grasps the idea of the peer. She elaborates on strengths that are important for the peer for her continuation of the task and adds explanations, thoughts, or examples to her statements and reasons why the strength is/strengths are important for the business idea.

\textit{Weaknesses:} The student thinks completely from the peer’s perspective and what would help him/her to further succeed with the task. The student explains the weakness in a very detailed manner and describes why the weakness is important to consider. He can also give counterarguments or ask questions to illustrate the weakness.

\textit{Suggestions for improvement: }The student suggests improvements as if he were in the peer’s position in creating the best possible solution. The student completes his suggestions with rich explanations on why he/she would do so and elaborates on the improvements in a very concrete and detailed way. Almost every suggestion is supported by further explanations.
 \\ \hline
 4 = fairly strong & The student thinks from the perspective of the peer. She elaborates in a way that serves the peer best to further establish the idea or activity. Each component is affirmed with further explanations. 
 
\textit{Strengths:} The student is able to recognize one or more strengths that are helpful for the peer to affirm their business idea and activity. He/She highlights contextual strengths rather than formal strengths. The student supports most statements with examples or further personal thoughts on the topic but might still be missing some reasonings.

\textit{Weaknesses: }The student thinks from the peer’s perspective and what would help him/her to further succeed with the task. This could be demonstrated by stating various questions and establishing further thoughts. The student explains the weakness and adds examples, but he/she is still missing some reasonings.

\textit{Suggestions for improvement: }The student suggests one or more improvements that are relevant for the further establishment of the activity and idea from the perspective of the peer. Most suggestions are written concretely and, if applicable, supported by examples. In most cases, the student explains why he/she suggests a change.
 \\ \hline
 3 = slightly weak / equal & The student tries to understand the perspective of the peer and adds further elaborations to her statements. However, her elaborations are not completely thought through and her feedback is missing some essential explanations, examples, or questions to make sure she understood everything correctly. 
 
\textit{Strengths:} The student mentions one or more strengths and explains some of them with minor explanations or examples on why it is seen as a strength. However, most strengths focus on formal aspects rather than contextual aspects.

\textit{Weaknesses:} The student states one or more weaknesses and explains some of them with minor explanations or examples. The student could also just state questions to illustrate the weakness in the peer’s business idea. Most weaknesses are not explained why they are such.

\textit{Suggestions from improvements: }The student suggests one or more improvements that are mostly relevant for the further establishment of the activity. The suggestions are written only on a high-level and most of them do not include further explanations or examples. The student explains only occasionally why he/she suggests a change or how it could be implemented.
 \\ \hline
 2 = very weak & The student does not try to understand the peer’s perspective. The student rather just tries to accomplish the task of giving feedback.
 
\textit{ Strengths:} The student mentions one or more strengths. They could be relevant for the peer. However, he does not add any further explanation or details.

\textit{Weaknesses: }The student states one or more weaknesses without explaining why they are seen as such. They could be relevant for the peer. However, the statements do not include any further elaboration on the mentioned weakness.

\textit{Suggestions for improvement:} The student suggests one or more improvements that could be relevant for the peer. However, the student does not explain why he/she suggests the change or how the suggestions for improvement could be implemented.
 \\ \hline
 1 = absolutely weak & The student’s feedback is very short and does not include the peer’s perspective. She does not add any further elaboration in her thoughts. 
 
\textit{Strengths:} The student only mentions one strength. This might not be relevant at all and lacks any further explanation, detail, or example.
 
\textit{Weakness:} The student only mentions one weakness. This might not be relevant at all and lacks any further explanation, detail, or example.

\textit{Suggestions for improvement: }The student only mentions one suggestion. The suggestion is not followed by any explanation or example and might not be relevant for the further revision of the peer.
 \\ \bottomrule
\end{tabular}
\caption{Detailed description of the cognitive empathy scores.}
\label{tab:detailedcognitivEmpathyScore}
\end{table*}

\begin{table*}[!htb]
\centering
\footnotesize
\begin{tabular}{c|p{12cm}}
\toprule
 Score & Description \\ \hline\hline
 5 = strong & The student is able to respond very emotionally to the peer’s work and fully represents the affectional state in her entire review. She illustrates this by writing in a very emotional and personal manner and expresses her feelings (positive or negative) throughout the review. Strong expressions include exclamation marks (!). Typical feedback in this category includes phrases such as “brilliant!”, “fantastic”, “excellent”, “I am totally on the same page as you”, “I am very convinced”, “Personally, I find this very important, too”, “I am very unsure”, “I find this critical”, “I am very sure you feel”, “This is compelling for me”, etc. 
 \\ \hline
 4 = fairly strong & The student is able to respond emotionally to the peer’s submitted activity with suitable emotions (positive or negative). She returns emotions in her feedback on various locations and expresses her feelings by using the personal pronoun (“I”, “You”). Some sentences might include exclamations marks (!). Typical feedback in this category includes phrases such as “I am excited”, “This is very good!”, “I am impressed by your idea”, “I feel concerned about”, “I find this very...”, “In my opinion”, “Unfortunately, I do not understand”, “I am very challenged by your submission”, “I am missing”, “You did a very good job”, etc. 
 \\ \hline
 3 = slightly weak / equal& The student occasionally includes emotions or personal emotional statements in the peer review. They could be quite strong. However, the student’s review is missing personal pronouns (“I”, “You”) and is mostly written in third person. Emotions can both be positive or negative. Negative emotions can be demonstrated with concern, missing understanding or insecurity (e. g., with modal verbs or words such as rather, perhaps). Typically, scale 3 includes phrases such as “it’s important”, “the idea is very good”, ”the idea is comprehensible”, “it would make sense”, “the task was done very nicely”, “It could probably be that”, etc. 
 \\ \hline
 2 = very weak& Mostly, the student does not respond emotionally to the peer’s work. Only very minor and weak emotions or personal emotional statements are integrated. The student writes mostly objectively (e.g., “Okay”, “This should be added”, “The task was done correctly”, etc.). In comparison to level 1, she might use modal verbs (might, could, etc.) or words to show insecurity in her feedback (rather, maybe, possibly). 
 \\ \hline
 1 = absolutely weak& The student does not respond emotionally to the peer’s work at all. She does not show her feelings towards the peer and writes objectively (e.g., no “I feel”, “personally” “I find this..” and no emotions, such as “good”, “great”, “fantastic”, “concerned”, etc.). Typical examples would be “Add a picture.” or “The value gap XY is missing.” 
 \\ \bottomrule
\end{tabular}
\caption{Detailed description of the emotional empathy scores.}
\label{tab:detailedemotionalEmpathyScore}
\end{table*}

\section{Disagreement Analysis}
\label{app:disagreement}

To analyze the disagreement between the three annotators, we created a confusion probability matrix (CPM) \cite{cinkova-etal-2012-managing} for the review components and the empathy level scores. A CPM contains the conditional probabilities that an annotator assigns to a certain category (column) given that another annotator has chosen the category in the row for a specific item. In contrast to traditional confusion matrices, a CPM also allows for the evaluation of confusions if more than two annotators are involved in an annotation study \cite{Stab2014AnnotatingEssays}.

\begin{table}[!htb]
\centering
\scriptsize
\begin{tabular}{c|ccccc}
\toprule
 & Suggestions & Weakness & Strength & None \\ \hline
 Suggestions & \textbf{0.6056} & 0.2970 & 0.0214 & 0.0759 \\ 
 Weakness & 0.2139 & \textbf{0.7009} & 0.0203 & 0.0648 \\ 
 Strength & 0.0264 & 0.0347 & \textbf{0.8340} & 0.1049 \\ 
 None & 0.0662 & 0.0784 & 0.0742 & \textbf{0.7812} \\ \bottomrule
\end{tabular}
\caption{CPM for review component annotations.}
\label{tab:confusion_probability_matrix_component}
\end{table}

Table \ref{tab:confusion_probability_matrix_component} shows that there is a broad agreement between the annotators in distinguishing between the different types of review components. The major disagreement is between suggestions and weaknesses, though with a score of 60\%, the agreement is still fairly high. Consequently, the annotation of review components in terms of strengths, weaknesses, and suggestions for improvements yields highly reliable results.

\begin{table}[!htb]
\begin{minipage}{0.48\textwidth}
\centering
\small
\begin{tabular}{c|cccccc}
\toprule
   & 1 & 2 & 3 & 4 & 5 \\ \hline
 1  & \textbf{.113} & .387 & .175 & .165 & .160 \\ 
 2  & .125 & \textbf{.266} & .362 & .211 & .035 \\ 
 3  & .025 & .159 & \textbf{.223} & .482 & .112\\ 
 4  & .014 & .054 & .283 & \textbf{.300} & .349 \\ 
 5  & .021 & .014 & .105 & .556 & \textbf{.303}\\ \bottomrule
\end{tabular}
\caption{CPM for cognitive empathy level annotations.}
\label{tab:confusion_probability_matrix_cognitive}
\end{minipage}
\hfill
\begin{minipage}{0.48\textwidth}
\centering
\small
\begin{tabular}{c|cccccc}
\toprule
   & 1 & 2 & 3 & 4 & 5 \\ \hline
 
 1 & \textbf{.106} & .459 & .286 & .086 & .063 \\ 
 2 & .154 & \textbf{.234} & .455 & .128 & .029 \\ 
 3 & .059 & .282 & \textbf{.350} & .240 & .068 \\ 
 4 & .026 & .115 & .347 & \textbf{.295} & .218 \\ 
 5 & .043 & .061 & .227 & .501 & \textbf{.168}\\ \bottomrule
\end{tabular}
\caption{CPM for emotional empathy level annotations.}
\label{tab:confusion_probability_matrix_emotional}
\end{minipage}
\end{table}

The CPMs for the empathy levels (see Tables \ref{tab:confusion_probability_matrix_cognitive} and \ref{tab:confusion_probability_matrix_emotional} reveal that there is a higher confusion between the scores assigned by the three reviewers, as compared to the annotation of the review components. However, when analyzed more closely, one can see that the scores mostly vary only within a small window of two or three neighboring scores. Therefore, we conclude that the annotation of cognitive and emotional empathy scores is reliably possible, too.

\begin{table}[!htb]
\small
\centering
\begin{tabular}{c|cccc}
\toprule
  & precision & recall & f1-score & support  \\ \hline
 non-empathic & 0.5746 & 0.5662 & 0.5704& 136\\ 
 empathic & 0.6364 & 0.5625 & 0.5972 & 112 \\ 
 neutral & 0.5240 & 0.5707& 0.5464 & 191 \\ 
 None & 0.9863 & 0.9729 & 0.9795 & 295\\ \hline
micro avg & 0.7322 & 0.7302 & 0.7482 & 734 \\ 
macro avg & 0.6803 & 0.6681 & 0.6734 & 734 \\ 
weighted avg & 0.7363 & 0.7302 & 0.7327 & 734 \\ 
samples avg & 0.7248 & 0.7302 & 0.7266 & 734 \\ \bottomrule
 
\end{tabular}
\caption{BERT model results for emotional empathy.}
\label{tab:resultsBertEmotional}
\end{table}

\begin{table}[!htb]
\small
\centering
\begin{tabular}{c|cccc}
\toprule
  & precision & recall & f1-score & support  \\ \hline
 non-empathic & 0.5739 & 0.3587 & 0.4415 & 184 \\ 
 empathic & 0.6434 & 0.5490 & 0.5925 & 286 \\ 
 neutral & 0.3062 & 0.4747 & 0.3723 & 198\\ 
 None & 0.9841 & 0.9802 & 0.9822 & 506\\ \hline
micro avg & 0.6949 & 0.6925 & 0.6937 & 1174 \\ 
macro avg & 0.6269 & 0.5907 & 0.5971 & 1174\\ 
weighted avg & 0.7225 & 0.6925 & 0.6996 & 1174 \\ 
samples avg & 0.6861 & 0.6925 & 0.6882 & 1174 \\ \bottomrule
 
\end{tabular}
\caption{BERT model results for cognitive empathy.}
\label{tab:resultsBertCognitive}
\end{table}

\begin{table}[!htb]
\small
\centering
\begin{tabular}{c|cccc}
\toprule
  & precision & recall & f1-score & support  \\ \hline
 non-empathic & 0.5739 & 0.3587 & 0.4415 & 184 \\ 
 neutral & 0.3062 & 0.4747 & 0.3723 & 198\\ 
 empathic & 0.6434 & 0.5490 & 0.5925 & 286 \\ 
 None & 0.9841 & 0.9802 & 0.9822 & 506\\ \hline
 f1 avg & 0.64 & 0.64 & 0.64 & 368\\ 
weighted avg & 0.73 & 0.73 & 0.73 & 368 \\ 

 \bottomrule
\end{tabular}
\caption{Results for the LSTM for emotional empathy.}
\label{tab:resultsLSTMEmotional}
\end{table}

\begin{table}[!htb]
\small
\centering
\begin{tabular}{c|cccc}
\toprule
  & precision & recall & f1-score & support  \\ \hline
 non-empathic & 0.74 & 0.28 & 0.40 & 83 \\ 
 neutral & 0.43 & 0.55 & 0.49 & 60\\ 
 empathic & 0.35 & 0.63 & 0.45 & 57 \\ 
 None & 0.99 & 0.94 & 0.97 & 168\\ \hline
 f1 avg & 0.63 & 0.60 & 0.58 & 368\\ 
weighted avg & 0.75 & 0.68 & 0.68 & 368\\ 
 \bottomrule
\end{tabular}
\caption{Results for the LSTM for cognitive empathy.}
\label{tab:resultsLSTMEmotional}
\end{table}

\section{Details on Application and Evaluation of Writing Support Tool}

To ensure the validity of our BERT model, we benchmarked against bidirectional Long-Short-Term-Memory-Conditional-Random-Fields classifiers (BiLSTM-CRF). In combination with the corresponding embeddings vocabulary (GloVe) \cite{Pennington2014GloVe:Representation}, our LSTM reached an unsatisfying f1-score of 61\% for detecting the emotional empathy level and 51\% for detecting the cognitive empathy level.

More information on the results of our BERT model and the LSTM for emotional and cognitive empathy detection can be found in the Tables \ref{tab:resultsBertEmotional}, \ref{tab:resultsBertCognitive}, \ref{tab:resultsLSTMEmotional}, and \ref{tab:resultsLSTMEmotional}.

In the post-survey, we measured perceived usefulness following the technology acceptance model \cite{Venkatesh2008TechnologyInterventions}. The items for the constructs were: "\textit{Imagine the tool was available in your next course, would you use it?}", "\textit{Assuming the learning tool would be available at a next course, I would plan to use it.}", or "\textit{Using the learning tool helps me to write more emotional and cognitive empathic reviews. }" Moreover, we asked the participants to judge their perceived empathy skill learning (PESL) by asking two items that cover cognitive and emotional empathy skills \cite{Spreng2009TheMeasures, Davis1983MeasuringApproach.}: \textit{“I assume that the tool would help me improve my ability to give appropriate emotional feedback.”} and \textit{“I assume that the tool would help me improve my ability to empathize with others when writing reviews.”} Finally, we surveyed the perceived feedback accuracy (PFA) \cite{Podsakoff1989EffectsPerformance} of both learning tools by asking three items: \textit{“The feedback I received reflected my true performance.”, “The tool accurately evaluated my performance.”}, and \textit{“The feedback I received from the tool was an accurate evaluation of my performance”.} All constructs were measured with a 1- to 7-point Likert scale (1: totally disagree to 7: totally agree, with 4 being a neutral statement). 

\begin{table} [!htb]
\small
\begin{tabular}{ p{2cm} | p{5cm} } 
\toprule
\textbf{Cluster} & \textbf{Feature} \\ \hline 
On empathy feedback reaction & \textit{"I think that this tool could help me not only to put myself in the position of a person in terms of content and make suggestions but also to communicate to them better"} \\ 
 \hline
On the feedback for skill learning & \textit{"The empathy feedback was clear and could be easily implemented. I had the feeling I learned something.Would use it again!"} \\ 
 \hline
 On cognitive and emotional empathy & \textit{"It was helpful that a distinction was made between the two categories of empathy. This again clearly showed me that I do not show emotional empathy enough. It was also useful that the tool said how to show emotional empathy (feelings when reading the business idea etc.)."} \\ 
 \hline
Improvements on feedback granularity & \textit{"It would be better if the feedback was more s elective or with detailed categories about empathy."}
 \\ 
 \hline
 Improvements on feedback recommendations & \textit{"Even more detailed information on how I can improve my empathy writing would be helpful, e.g., with review examples."}
 \\ 
 \bottomrule
\end{tabular}
\caption{Representative examples of qualitative user responses after the usage of our empathy support tool.}
 \label{tab:qualitativeuser}
\end{table}

\end{document}